\documentclass[10pt,twocolumn,letterpaper]{article}

\usepackage[pagenumbers]{cvpr} %

\usepackage{graphicx}
\usepackage{amsmath}
\usepackage{amssymb}
\usepackage{booktabs}

\usepackage{mathtools}
\usepackage{iac_pkg}
\usepackage{lipsum}
\usepackage{multirow}
\usepackage{tabularx, booktabs}
\usepackage{color}
\usepackage{xspace}
\usepackage{overpic}
\usepackage{xcolor}
\definecolor{lightred}{RGB}{235, 148, 149}
\definecolor{lightblue}{RGB}{173, 216, 229}
\definecolor{citecolor}{RGB}{34,139,34}
\usepackage[pagebackref=true,breaklinks=true,letterpaper=true,citecolor=citecolor,colorlinks,bookmarks=false]{hyperref}

\usepackage[capitalize]{cleveref}
\crefname{section}{Sec.}{Secs.}
\Crefname{section}{Section}{Sections}
\Crefname{table}{Table}{Tables}
\crefname{table}{Tab.}{Tabs.}

\def\cvprPaperID{*****} %
\def\confName{CVPR}
\def\confYear{2023}

\begin{document}

\title{Exploring the Connection between Robust and Generative Models}

\author{Senad Beadini\\
Computer Science Department\\
Sapienza, University of Rome\\
{\tt\small beadini.senad@gmail.com}
\and
Iacopo Masi\\
Computer Science Department\\
Sapienza, University of Rome\\
{\tt\small masi@di.uniroma1.it}
}

\maketitle

\begin{abstract}
We offer a study that connects robust discriminative classifiers trained with adversarial training (AT) with generative modeling in the form of Energy-based Models (EBM). We do so by decomposing the loss of a discriminative classifier and showing that the discriminative model is also aware of the input data density. Though a common assumption is that adversarial points leave the manifold of the input data, our study finds out that, surprisingly, untargeted adversarial points in the input space are very likely under the generative model hidden inside the discriminative classifier---have low energy in the EBM. We present two evidence: untargeted attacks are even more likely than the natural data and their likelihood increases as the attack strength increases. This allows us to easily detect them and craft a novel attack called High-Energy PGD that fools the classifier yet has energy similar to the data set. The code is available at the link above.%
\end{abstract}

\section{Introduction}
Although the attention to machine learning robustness and its security implications became suddenly of interest, the idea of \emph{robust} of machine learning (ML) is not new. Pioneering articles back in 2004 raised awareness that the community naively assumed that the data generation process in data mining is independent of the system responses~\cite{dalvi2004adversarial}.
A few years later, \cite{barreno2010security} provided the first consolidated perspective of adversarial ML and the need for a change in how ML is thought, conceived, and applied. 
Standard, non-robust ML has three limits in the adversarial setting: (1) the first limitation is more proper of the generalization ability of ML systems. These models operate under the assumption that the testing data is sampled from the same unknown generation process that created the data--i.i.d. assumption, independent and identically distributed. This is a too strict assumption when actual input data can be sampled from another generation process--o.o.d., out of distribution; (2) more importantly, the generation process is assumed to be disconnected from the system capabilities of taking decisions (3) minimization of the empirical risk does not guarantee to be resilient to an adversary.
To address those issues, the formalization of minimizing the maximum risk~\cite{wald1945statistical} by A. Wald of 1944 came at hand to cast the empirical risk minimization into \emph{adversarial risk minimization}~\cite{goodfellow2014explaining,madry2017towards}, following a proper definition. %

\begin{figure}[tb]
    \centering
 \begin{overpic}[keepaspectratio=true,width=.9\linewidth]{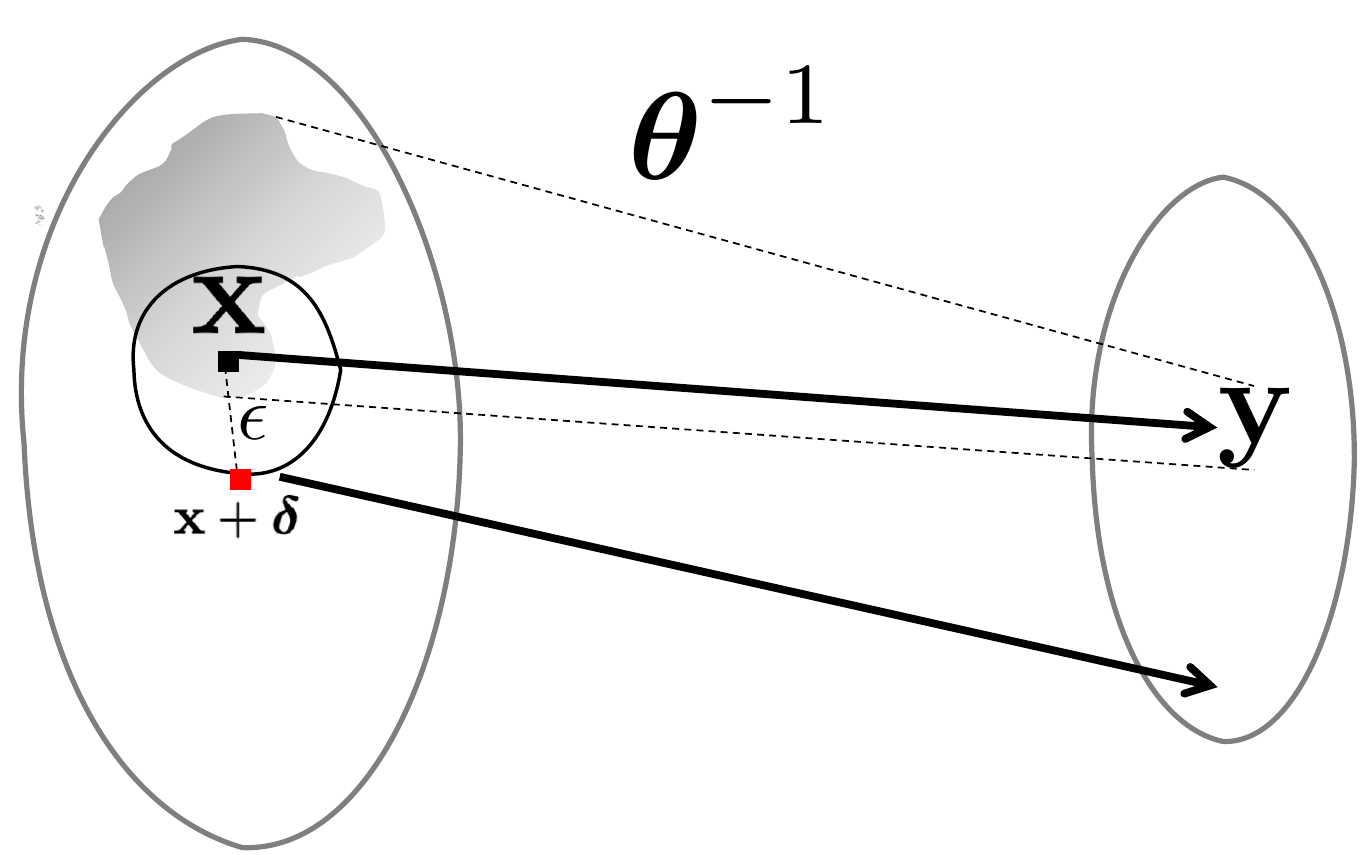}
    \end{overpic}
    \caption{An input point $\bx$ on the left is mapped to a latent code $\by$ to solve a generic task (i.e. classification). When we invert the latent code $\by$ and we go back in the input space, we have a full set of solutions. Adversarial examples are points \emph{not} in that set and also very close to $\bx$ yet mapping to another latent code, far from $\by$: for small variations of the input, we have large variations of the output. Adversarial training fixes these points, making the model more ``aware'' of the distribution of the input data.}
    \label{fig:inversion}
    \vspace{-15pt}
\end{figure}

In this paper, following the interpretation in~\cite{grathwohl2019your} of a discriminative classifier $p_{\net}(y|\mbf{x})$ implemented by a Convolutional Deep Neural Network (DNN) for multi-class classification, we offer a new view on adversarial perturbations connecting them with generative modeling. Following~\cite{grathwohl2019your,zhu2021towards}, we analyze a robust model under the lens of Energy-based Models (EBM), and we decompose the cross-entropy loss of $p_{\net}(y|\mbf{x})$ to show that it contains the notion of the joint distribution of the data and the labels $p_{\net}(\bx, y)$ along with the distribution of the data itself $p_{\net}(\bx)$. Although adversarial perturbations are known as input points that switch the decision boundary---thus affecting $p_{\net}(y|\mbf{x})$---we show that there is a strong dependency also with $p_{\net}(\bx)$. In the scenario of untargeted attacks, we show that an increase in the attack strength implies a significant increase of $p_{\net}(\bx)$. The increase is so substantial that the attacks' energy surpasses the magnitude of the energy of natural data. Under the lens of generative modeling via EBM, we hope to shed some light on a few phenomena that emerge from robust classifiers, such as: 1) robust models are naturally better calibrated than standard classifiers, 2) robust models offer gradients of the loss wrt to the input with more structure, and better ``explain'' the data 3) the trade-off between robust and natural accuracy. \cref{fig:inversion} shows the idea of training a robust model: in the standard case, when we train a discriminative classifier, we model $p_{\net}(y|\mbf{x})$, and thus we seek for a function $\net$ that maps the fixed input point $\bx$ to a latent 
code $\by$ that is useful for the task, for instance, minimize the cross-entropy loss. A discriminative 
model ignores the input data distribution, thereby adjusting its parameters to separate the data. 
Furthermore, the input space is usually larger than the embedding space, and the model has to build invariance to nuances of the training data to achieve classification. %
What we \emph{do not enforce}, instead, is that the inverse of $\net$ has to fall within a boundary of the 
input data. In other words, when we train the discriminative model, the function is not bijective, and inverting the function gives as output a set. Adversarial samples are points in the input space very close to $\bx$ yet map to a different latent code far from 
$\by$. Adversarial Training (AT)~\cite{goodfellow2014explaining,madry2017towards} restores the ``right'' connection. %

In this paper we make the following contributions:

$\diamond$ Viewing a discriminative model under the lens of an EBM, we show that the more we increase the strength of an untargeted attack, the more $p_{\net}(\bx)$ increases, even more than the points naturally present in the dataset.

$\diamond$  We show that the energy of an input point $\bx$ can be used as a strong detector for adversarial data. The more the attack increases its strength, the easier to detect. We do so by performing \emph{no training} of the detector except choosing a threshold.

$\diamond$ We offer a way to bypass the aforementioned detector demonstrating that is possible to design an adversarial perturbation that crosses the decision boundary yet its energy is kept similar to those of natural data.

$\diamond$ Finally, we emphasize that a robust or quasi-robust model can be effectively employed to enhance image synthesis in generative AI, supported by a recent paper by~\cite{rouhsedaghat2022magic}, where image synthesis is performed by inverting a quasi-robust model.

\section{Prior Work}

\minisection{Robust and Generative Models} We believe the first connection between robust and generative models has been done in the paper by Grathwoh \etal \cite{grathwohl2019your}, where for the first time, a discriminative DNN ending with the softmax classifier is interpreted as an EBM. \cite{grathwohl2019your} proposes JEM to train the DNN in a hybrid way---to be both discriminative and generative. They also show that hybrid training somehow slightly improves robustness to adversarial attacks. JEM was extended in~\cite{yang2021jem++} to stabilize and speed up the training.
Later on,~\cite{zhu2021towards} established the first connection on how adversarial training (AT) and EBM bend the energy function in a different way yet using a similar contrastive methodology. Very recently~\cite{wang2022a} moved in the same direction of demystifying the generative capabilities of robust models developing a unified probabilistic framework, and working also in unsupervised settings. Finally,~\cite {korst2022adversarial} showed that AT improves Joint Energy-Based Generative Modelling, while incorporating into the modeling a sharpness-aware minimization (SAM) procedure~\cite{foret2021sharpnessaware}.

\minisection{Synthesis with a robust model} At the best of our knowledge~\cite{santurkar2019singlerobust} is the first paper to employ a \emph{robust} classifier for synthesis since it is shown that robust models attain input gradients that better correlated with ``human perception''~\cite{aggarwal2020benefits,kaur2019perceptually}. The community argued in~\cite{terzi2020adversarial} that AT renders the discriminative model invertible, while ~\cite{kim2019bridging} believes that AT yields gradients closer to the image manifold, which is in line with the connection between robust and generative model. 
The generative capabilities of robust models have been used by~\cite{rojas2021inverting} for solving inverse problems and for controllable image synthesis~\cite{rouhsedaghat2022magic}.

\section{Preliminaries}
In~\cref{subsec:adv}, we will briefly review the settings of adversarial attacks in a white box scenario, while~\cref{subsec:ebm} we examine data density modeling of the input data using an Energy-based Model (EBM). 

\subsection{Adversarial Perturbations}\label{subsec:adv}
A DNN $\net : \mathbb{R}^{X} \rightarrow \mathbb{R}^{Y}$ is a universal approximator for generic functions~\cite{csaji2001approximation} that maps an input data $\bx$ into another meaningful code $\by$ using a loss function $\mathcal{L}$. In machine vision, usually $X$ is high dimensional, with $X \gg Y$, and $\mathcal{L}$ guides $\net$ in solving the task at hand. The definition of adversary specifies that, though $\net$ may perform well on the task yet $\net$ is brittle since is possible to obtain a slightly perturbed instance of $\bx^{\star}$ that fools $\net$ on the same task such that $\bx^{\star} \approx \bx$. 
The gist of all the attacks follows an inverse optimization process in which the attacker optimizes an adversarial point $\bx^{\star} \in \mathbb{R}^{X}$ in the input space so that the model ascends the loss in the output space (i.e. untargeted attack) or makes a confident response of an erroneous label, decided a priori by the attacker (i.e. targeted attack). 
The attacker often yields a constraint on how much information needs to be changed in the input. %
In most of the current attacks, the bound on the information is implemented with pixel-wise $L_p$ norm as $\bdelta\doteq \bx^{\star} - \bx$ and $||\bdelta||_p \leq \epsilon$.  All the attacks follow the above procedure with some differences on how many steps are taken to ascend the loss: one for Fast-Gradient Sign Method (FGSM) with its variants~\cite{goodfellow2014explaining,dong2018boosting}, multi-steps with projections for Projected Gradient Descent (PGD)~\cite{madry2017towards} or in how the loss is designed, e.g. Carlini \& Wagner (CW) attack. 

Attacks can be crafted in targeted or untargeted settings even in the physical world~\cite{kurakin2016adversarial} and can transfer between different models. %
Whilst attack literature is solid, defenses have several limitations~\cite{tramer2020adaptive}: despite being computationally expensive, the only defense to withstand is Adversarial Training (AT)~\cite{goodfellow2014explaining,madry2017towards}.
Though in this paper we deal mainly with the image domain, adversarial perturbations are known in other domains (Natural Language Processing and tabular data) as counterfactual explanations~\cite{madaan2021generate}. Below we briefly review the two main methods to generate perturbations of the input given a DNN.

\minisection{Fast Gradient Sign Method} FGSM \cite{goodfellow2014explaining} is a $L_\infty$-norm adversarial attack and it is described in the following formula below:
\begin{equation}
\!\bx^* = \bx + \epsilon \;\operatorname{sign}\Big( \; \nabla_{\bx} \big[\mathcal{L}(\net(\bx), \;y)\big] \;\Big) 
\label{FGSM}
\end{equation}

where $\operatorname{sign}$ returns the sign of its argument.

\minisection{Projected Gradient Descent}
PGD can be seen as an iterative generalization of the FGSM algorithm \cite{madry2017towards}. The iterative procedure is shown below:
\begin{equation}
\!\bx^* = \operatorname{clip}_\epsilon \;\Big(\bx^* + \alpha \; \operatorname{sign}\Big(\; \nabla_{\bx^*} \mathcal{L}(\net(\bx^*), \;y)\;\Big)\Big )
\end{equation}

where $\operatorname{clip}$ denotes the function that projects its argument to the surface of $\bx$'s neighbor $\epsilon$-ball while $\alpha$ is the step size. This algorithm is able to produce strong adversarial data that have a high probability to fool classifiers and commonly is used as a benchmark.
Additional variants of the PGD attack, such as APGD~\cite{croce2020reliable}, are accessible and can improve its reliability.

\subsection{Energy-Based Model (EBM)}\label{subsec:ebm}

EBMs \cite{lecun2006tutorial} are based on the idea that any probability density $p_{\net}(\bx)$ for $\bx \in \mathbb{R}^X$ can be expressed via a Boltzmann distribution as: 
\begin{equation}
p_{\net}(\bx) = \frac{\exp{(-E_{\net}(\bx))}}{Z(\net)}
\label{EBM1}
\end{equation}
where $E_{\net}(\bx)$ is called energy, modeled as a neural network, that maps each input $\bx$ to a scalar. $Z(\net)$ is the normalizing constant, known as the partition function, such that $p_{\net}(\bx)$ is a valid probability density function;
the challenge for training EBMs is approximating the constant $Z(\net)$. 
Training an EBM is performed via maximum likelihood estimation by minimizing the negative log-likelihood of the data: 
\begin{equation}
    \mathcal{L}_{ML}( \net ) = \EX_{\bx \sim P_D} \Big[ - \log{p_{\net}(\bx)} \Big]
    \label{emb-ml}
\end{equation}

Nevertheless, the latter is not effortless and several sampling methods have been designed to approximate it efficiently ~\cite{song2021train}. Specifically, it is known that the derivative of \cref{emb-ml} w.r.t $\net$ is:
\begin{equation} \centering
\resizebox{\hsize}{!}{
    $\nabla_{\net} \mathcal{L}_{ML}(\net) = \EX_{\bx^+ \sim P_D} \Big[ \nabla_{\net} E_{\net}(\bx^+) \Big] -  \EX_{{\bx^-} \sim p_{\net }} \Big[ \nabla_{\net} E_{\net}(\bx^-) \Big]$
    }
    \label{grad-loss}
\end{equation}

Unfortunately, for EBMs we can not sample from $p_{\net}$, therefore several MCMC methods have been proposed to sample from that density function ~\cite{song2021train}.
For instance, in ~\cite{grathwohl2019your} Stochastic Gradient Langevin Dynamics (SGLD) has been used to train EBMs using gradient knowledge. SGLD sample with the following procedure:
\begin{equation} \centering
\begin{split}
    \bx_{0} \sim p_0(\bx) \;\;\; , \;\; \bx_{t+1} = \bx_t - \frac{\alpha}{2}  \frac{\partial E_{\net}(\bx)}{\partial \net} + \alpha \epsilon
    \label{eq:EBM3}
\end{split}
\end{equation}

where $p_0(\bx)$ is typically a Uniform distribution and $\epsilon \sim \mathcal{N}(0,1)$.
Thus, we can summarize that the training of EBMs consists of two steps: generating approximate data samples from $p_{\net}$ using MCMC methods and optimizing the model parameters to increase the energy of these samples and decreasing the energy of the real samples via gradient descent. 

\minisection{DNN as an EBM}
Recently, Grathwohl \etal \cite{grathwohl2019your} made a connection between DNN classifiers and EBMs. They show that an EBM is hidden inside a standard classifier and we can re-interpret the logits of a model  $\net$ to build an EBM to compute $p(\bx,y)$ and $p(\bx)$: 
\begin{equation}
p_{\net}(\bx, y) \; = \; \frac{\exp{(\net}(\bx)\;[y])}{Z(\net)}
\label{EBM4}
\end{equation}

where $Z(\net)$ is the unknown normalizing constant, $\net(\bx) [k]$ is the k-th index of logits $\net(\bx)$, and 
$E_{\net}(\bx,y) = -\net(\bx) [y]$ is the energy function. Following \cite{grathwohl2019your} we can compute $p_{\net}(y \; | \bx)$ and $p_{\net}(\bx)$ as:
\begin{equation*}
\begin{split}
& p_{\net}(y | \bx) = \frac{p_{\net}(\bx, y)} {p_{\net}(\bx)} = \frac{\exp{\big( - E_{\net}(\bx,y)\big)}} {\sum_{k=1}^{K} \exp{\big(\net(\bx) [k]\big)} } \\  &
p_{\net}(\bx) = \sum_{k=1}^{K} p_{\net}(\bx, y=k) = \frac{\sum_{k=1}^{K}{\exp{(\net(\bx)[k])}}} {Z(\net)}
\end{split}
\label{EBM5}
\end{equation*}
with $K$ the number of classes.
Notice that, for any classifier $\net$, the energy of a data point $\bx$ is:
\begin{equation}
E_{\net}(\bx) \; = \;  - \log \; \sum_{k=1}^{K} \exp{\big(\net(\bx)[k]\big)}
\label{EBM6}
\end{equation}
Indeed a data point with a high probability of occurrence---high $p_{\net}(\bx)$---is equivalent to having low energy.
\section{Method}

\subsection{Untargeted attacks have low $E_{\net}(\bx)$}

\begin{figure}[tb]
\centering
\subfloat[]{\label{a}\includegraphics[width=0.5\columnwidth,height=3cm]{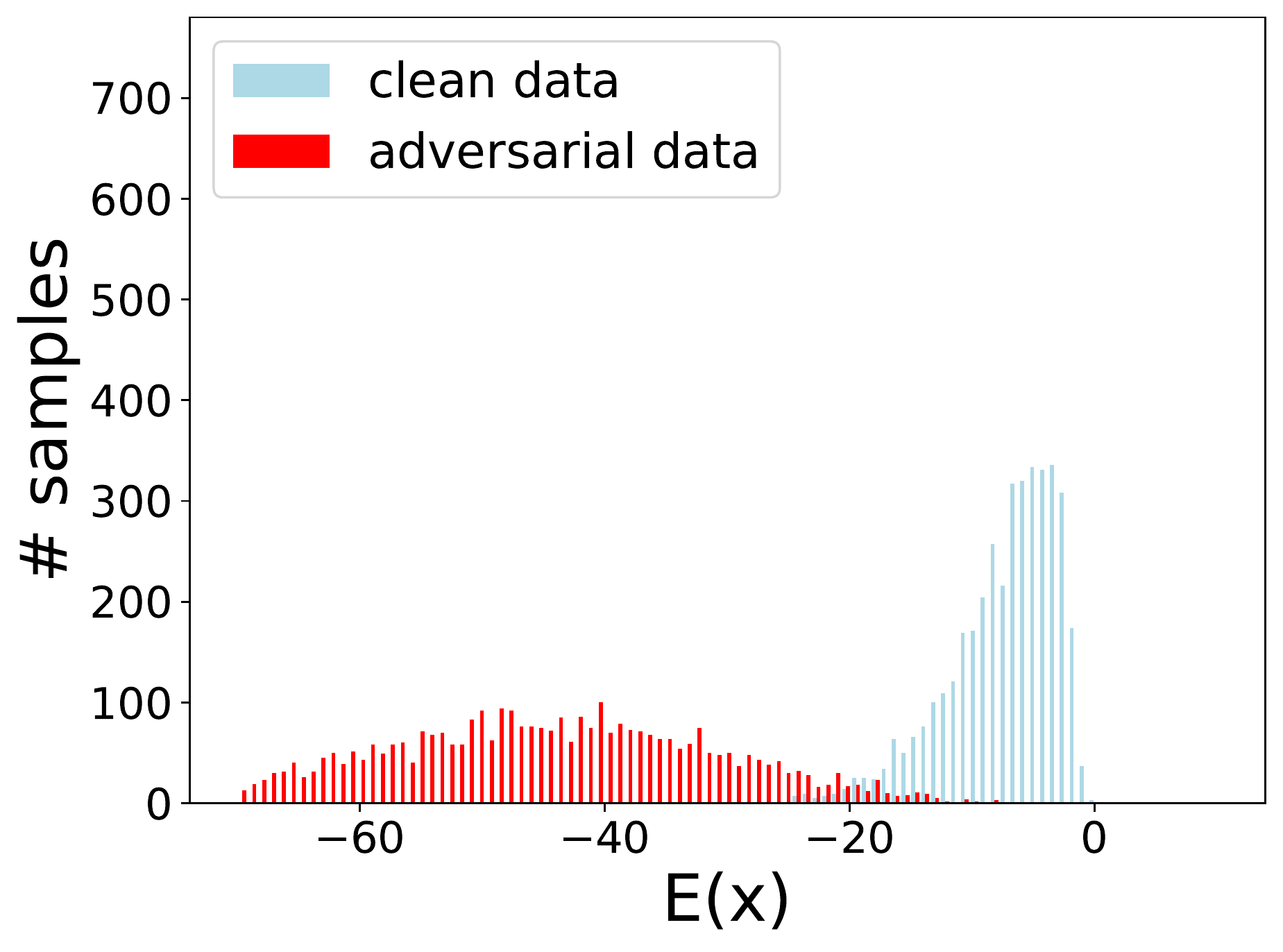}}
\subfloat[]{\label{c}\includegraphics[width=0.5\columnwidth,height=3cm]{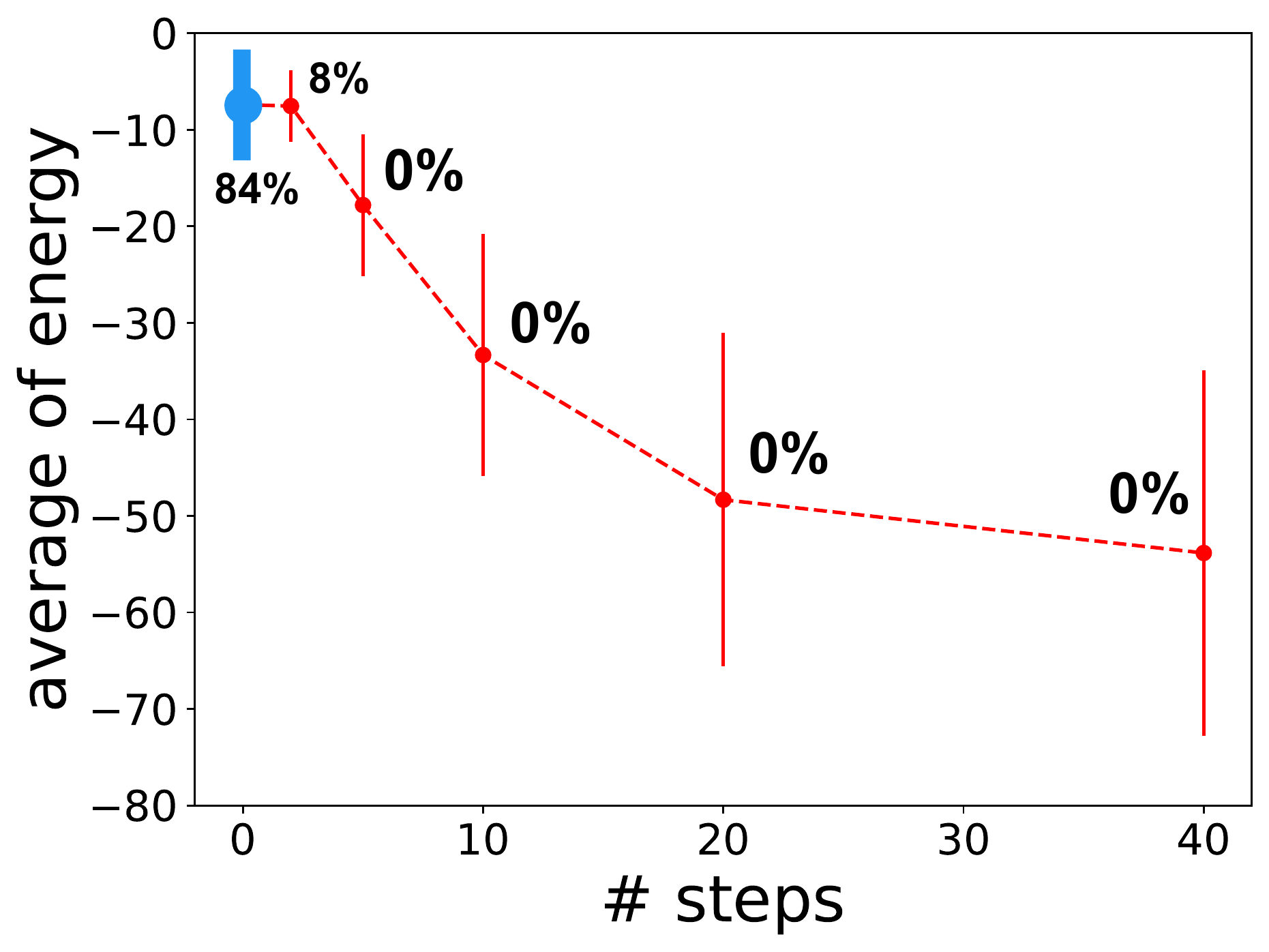}}\par
\caption{\textbf{(a)} Energy distribution $E_{\net}(\bx)$ for PGD-based adversarial and natural data on the test set of imagenette, given a standard non-robust DNN. Notice how the distributions are almost separated.
\textbf{(b)} Each point indicates the mean absolute value of the energy and its spread taken at different PGD ``steps'' (increasing strength). Above each point we reported the accuracy. \textcolor{lightred}{\rule{0.4cm}{0.25cm}} indicates adv. while \textcolor{lightblue}{\rule{0.4cm}{0.25cm}} natural data.}
\label{fig:result3}
\end{figure}

Analyzing the energy $E_{\net}(\bx)$ gives us the opportunity to examine and study the generative model inside classifiers. To better understand the characteristics of adversarial examples, we performed an energy analysis of DNNs for both adversarial examples generated using the untargeted PGD attack and natural data. Our results indicate a substantial difference in energy between adversarial and natural data, with adversarial points having much lower energy than natural points. 
These findings are exhibited in \cref{fig:result3}, which displays the energy for two different models on diverse datasets. In ~\cref{fig:result3}(a), the \#steps (iterations) used is 40 in PGD and $\epsilon$ is always 8/255.
Indeed, our analysis reveals an intriguing dependency between the energy function $E_{\net}(\bx)$ and the ``strength'' of an untargeted adversarial attack. It appears that the energy of adversarial data decreases as the strength of the attack increases. In our case, we define the attack strength as the discrete value of the number of steps of a PGD attack.
Considering these results and reinterpreting them using the EBM framework, it can be inferred the following. Beyond the classic idea that adversarial attacks cross the decision boundary, we show that a DNN tends to ``believe'' that adversarial data are highly likely under the hidden generative model $p_{\net}(\bx)$. More surprisingly, they are even more likely than the natural data itself.

\subsection{Detecting Adversarial Noise with the Energy Function} \label{sec:detector}
Based on the previous observations we propose the construction of a very simple detection algorithm for catching adversarial data using the energy function as a 1D discriminant. This idea has already been applied in ~\cite{liu2020energy} to detect out-of-distribution (OOD) data; however, there is a key difference from~\cite{liu2020energy} and us: while the OOD data tend to move to higher values of $E_{\net}(\bx)$, in our case, we have the opposite; additionally, to the best of our knowledge, this technique was not yet applied to adversarial example detection.
Our strategy detects PGD adversarial perturbations at inference time in a DNN equipped with a softmax classifier, without any additional computation overhead and any additional parameters. Firstly, the approach estimates a threshold $t$ in the domain of $E_{\net}(\bx)$ using a validation set that balances the true positive rate (TPR) and false negative rate (FNR) based on a chosen metric. Subsequently, it uses this threshold to perform detection: 
\begin{equation}
    G(\net, \bx) =
    \begin{cases}
      1 & \text{if \; $- \log \; \sum_{k=1}^{K} \exp{\big(\net(\bx)[k]\big)} \leq \; t$}\\
      0 & \text{otherwise}
    \end{cases}   
\end{equation}

\subsection{Attacking The Energy Detector}
The development of strong adversarial detectors is a significant research area, as such detectors typically depend on specific types of attacks. Although the energy approach is effective for identifying adversarial data with low energy, such as those generated via PGD, a crucial question is whether it is possible to generate perturbations that can mislead the aforementioned method. That means whether it is possible to produce adversarial examples that own energy comparable to natural data.

\minisection{High-Energy PGD} In order to generate adversarial examples with similar energy as natural data, we propose to introduce a regularization term during the optimization step to force the creation of perturbations that lead to adversarial data with higher energy. The regularizer will remove the energy gap between natural and adversarial data. Specifically, we define the following optimization problem:
\begin{equation}
    \arg \max_{\bdelta} \Big[ \mathcal{L} \big( \;\net(\bx + \bdelta),y \big) + \lambda \; E_{\net}(\bx+\bdelta) \Big]
\end{equation}

where $\mathcal{L}$ is the loss function, like cross-entropy and $\lambda$ is a hyper-parameter that controls the strength of the regularization. \cref{eq:HEPGD} describes the High-Energy PGD attack (HE-PGD) implemented as:

\begin{equation}
\!\bx^* = \operatorname{clip}_\epsilon \Big[\bx^* + \alpha  \operatorname{sign}\Big[ \nabla_{\bx^*} \mathcal{L}\big(\net(\bx^*), y \big)+\lambda E_{\net}(\bx^*) \Big]\Big]%
    \label{eq:HEPGD}
\end{equation}

where the initial $\bx_0^*$ = $\bx + \bdelta$ with $\bdelta \sim$ Uniform(-$\epsilon$,$\;$ $\epsilon$).\\
The choice of $\lambda$ is crucial and can significantly affect optimization and thus the aggressiveness of the attack. The empirical value of the $\lambda$ parameter is described in~\cref{sec:exp}.

\section{Experimental Results} \label{sec:exp}

In this section, we describe our experiments and confirm the efficacy of the energy function as a detection method for PGD-based attacks. Then, we investigate a connection between robust and generative models.

\subsection{Detection Results}

We use the imagenette \cite{url2}, CIFAR-10 \cite{krizhevsky2009learning}, and CIFAR-100 \cite{krizhevsky2009learning} datasets as benchmarks.
All attacks except HE-PGD are implemented using torchattacks \cite{kim2020torchattacks}. HE-PGD is implemented directly in PyTorch with a $\lambda$ parameter of $1.2$.

\minisection{Evaluation Metrics} We measure the following metrics: 1) detection rate (DR); 2) false positive rate (FPR).
We utilize the G-mean metric to estimate the threshold; G-mean is defined as the squared root of the product of sensitivity and specificity.
\cref{tab:performance} shows the detector's performance on diverse attack settings. Energy as a detector achieves larger DR for PGD-based perturbations on CIFAR-10, also scale to imagenette~\cite{url2} where the input space is much larger than CIFAR, and works for unseen attacks such as APGD~\cite{croce2020reliable}.

\begin{table}[h]
    \centering
    \begin{tabular}{c|c|c|c|l} 
    \hline
    Dataset & Defense & Attack & DR & FPR  \\
    \hline
    \multirow{2}{4em}{imagenette\\~\cite{url2}} & \multirow{2}{4em}{\tbf{Energy} (ResNet10)} & PGD~(8) & 98.24 & 1.37 \\
    & & PGD~(16) & 99.6 & 0.00 \\
    \hline
    \multirow{5}{4em}{CIFAR-10~\cite{krizhevsky2009learning}} & \multirow{2}{4em}{\tbf{Energy} (ResNet10)} & PGD~(8) & 98.38 & 1.62 \\
    & & APGD~(8) &  85.45 & 1.19 \\
    & KD+BU\cite{feinman2017detecting} & PGD~(8) & 92.27 & 0.96 \\
    & LID \cite{ma2018characterizing} & PGD~(8) & 94.39 & 1.81 \\
    \hline
    \end{tabular}
    \caption{Attack detection with energy function. KD+BU and LID results taken from \cite{aldahdooh2022adversarial}. All numbers are percentages. Attacks are expressed with the $(\epsilon)$ budget.\vspace{-10pt}}
    \label{tab:performance}
\end{table}

\subsection{HE-PGD Results}
\begin{figure}[h]
\centering
\subfloat[]{\label{a}\includegraphics[width=0.5\columnwidth,height=2.85cm]{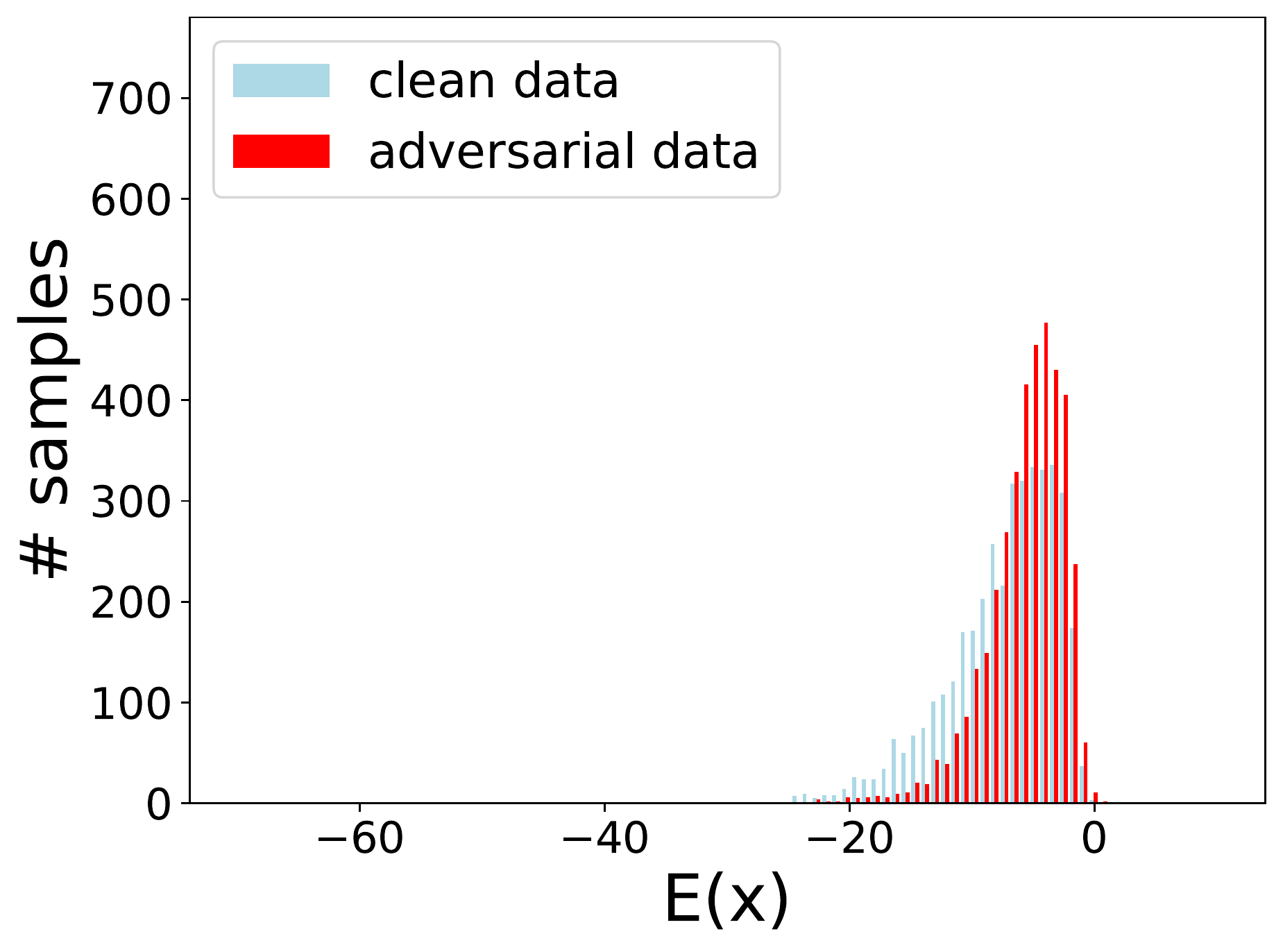}}
\subfloat[]{\label{c}\includegraphics[width=0.5\columnwidth,height=2.85cm]{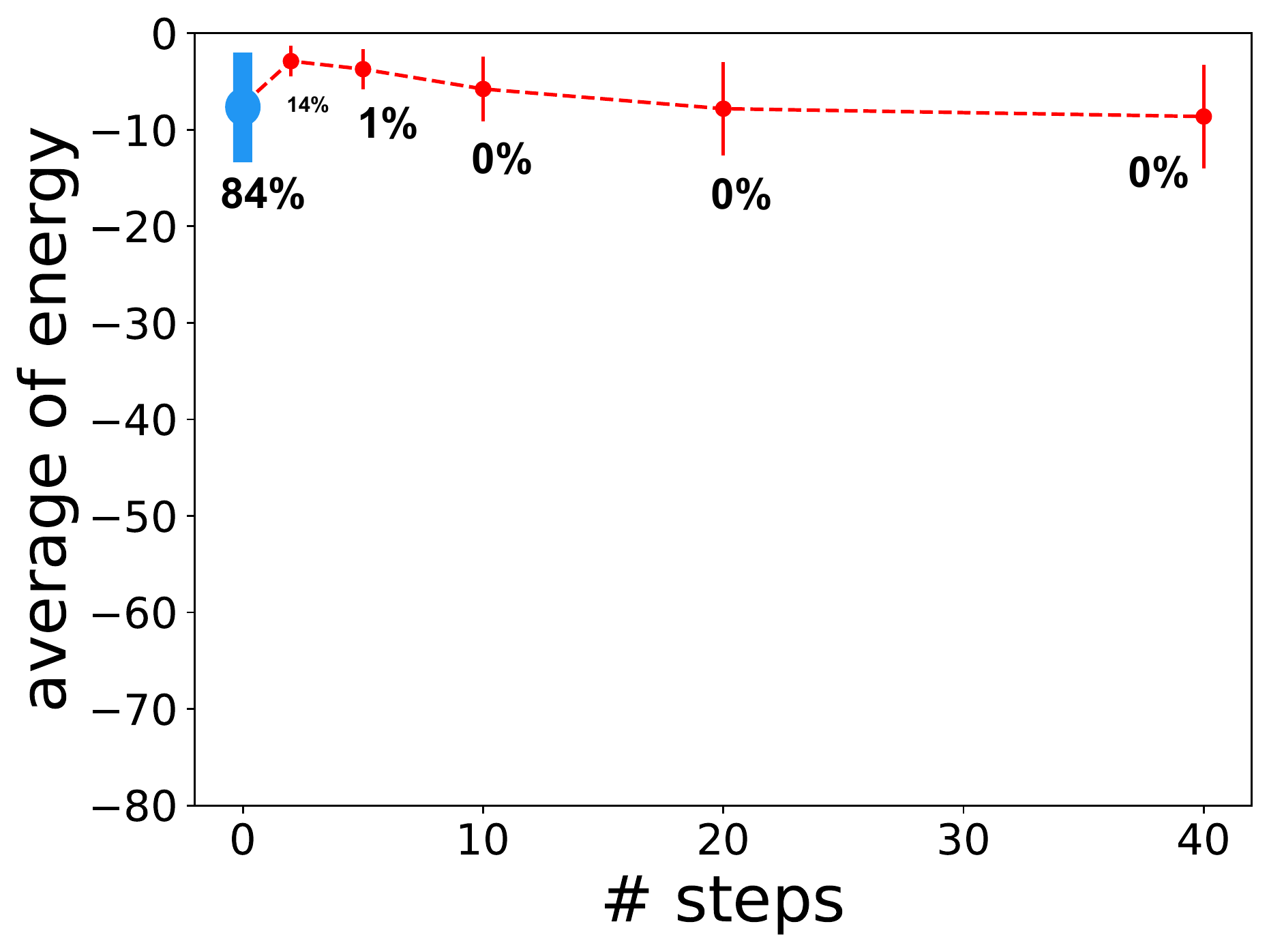}}\par
\caption{\textbf{(a)} On imagenette, energy distribution for HE-PGD adversarial and natural data. The two distributions are overlapped, which makes the energy detector worthless. \textbf{(b)} HE-PGD enforces invariance of $E_{\net}(\bx)$ wrt to the \#steps yet still pushes the accuracy to 0\% after a few steps. \textcolor{lightred}{\rule{0.4cm}{0.25cm}} indicates adv. while \textcolor{lightblue}{\rule{0.4cm}{0.25cm}} natural data.}
\label{fig:result1}
\end{figure}

Our experiments in \cref{fig:result1} show that HE-PGD is at least as effective as conventional PGD while it keeps the energy of the adversarial points similar to the natural data. Indeed in \cref{fig:result1}(a) the two distributions now largely overlap since the adversarial samples generated by HE-PGD have higher energy than those produced by PGD and are better aligned with the energy of the natural data. Consequently, when the value of $\lambda$ is chosen appropriately, HE-PGD can bypass the energy-based detector, presented in~\cref{sec:detector}. The stepsize of all attacks is fixed at 1/255.

From \cref{fig:result1}(b), we note that, unlike before, $E_{\net}(\bx)$ is approximately invariant to the number of steps (\#steps) in projected gradient descent. At the same time, the classifier accuracy still drops to zero for higher iterations yet it does so slightly more gradually than before.

\subsection{Synthesis with a Robust Model} 
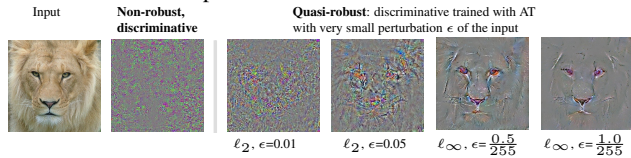
\begin{figure}[tb]
\centering
 \begin{overpic}[width=\linewidth]{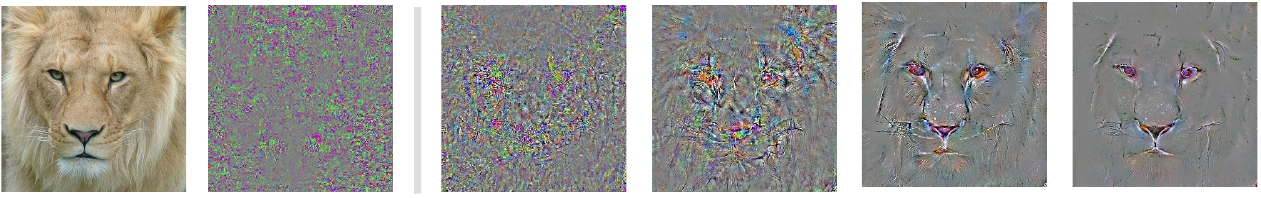}
    \put(4,19){\tiny{Input}}
    \put(17.5,19){\tiny{\tbf{Non-robust,}}}
    \put(17.5,16.5){\tiny{\tbf{discriminative}}}
    \put(45.5,19){\tiny{\tbf{Quasi-robust}: discriminative trained with AT}}
    \put(45.5,16.5){\tiny{with very small perturbation $\epsilon$ of the input}}
    \put(36,-2){\tiny{$\ell_{2}$, $\epsilon$=0.01}}
    \put(53.5,-2){\tiny{$\ell_{2}$, $\epsilon$=0.05}}
    \put(68.5,-2){\tiny{$\ell_\infty$, $\epsilon$=$\frac{0.5}{255}$}}
    \put(85.5,-2){\tiny{$\ell_\infty$, $\epsilon$=$\frac{1.0}{255}$}}
 \end{overpic}
\caption{A regular discriminative model shows gradients of the loss respect to the input with sparse activations compared to robust models while these latter are more meaningful which better motivate them for synthesis and connect them to generative models.\vspace{-10pt}}
\label{fig:gradients}
\end{figure}

The evidence that robust models have input gradients aligned with semantic properties of the object, as shown in~\cref{fig:gradients}, can be explained under the lens of generative modeling. Aggregating gradients of the loss over the input can be seen as a way to sample points from the generative model hidden inside the discriminative classifier.
What AT may be doing is rendering it a hybrid model that becomes generative by estimating gradients of the data distribution as a score-matching model~\cite{song2019generative}. There is also a strong parallelism between adversarial attacks and sampling from score-matching using SGLD---\cref{eq:EBM3}. Recently~\cite{rouhsedaghat2022magic} showed that is possible to invert a quasi-robust model to perform convincing image synthesis. %
A quasi-robust model is a model which is non-robust from the point of view of security yet it inherits generative capabilities and keeps its well-performing accuracy. To generate the samples,~\cite{rouhsedaghat2022magic} takes steps in the direction of the gradients that better carries information of class conditional data distribution $p(\bx|y)$.

\section{Conclusions} By viewing a robust discriminative model as an EBM, we were able to detect adversarial example effectively and demonstrate that there is the possibility of bypassing the detector introducing a new attacked called High-Energy PGD. We believe that exploring the connection between robust and generative modeling may shed some light on interesting properties of robust classifiers.
We experimentally showed that untargeted adversarial examples (PGD) have lower energy than natural data. Then, under this validation, we provided a detector for recognizing PGD attacks using the energy function of the generative hidden model inside classifiers. Finally, we indicate that a robust model or quasi-robust model can be employed as an image generator.

{\small \minisection{Acknowledgment} The authors thank Dr. M Rouhsedaghat for providing the figure of the gradients. This project is partially supported by Sapienza research project ``Prebunking: predicting and mitigating coordinated inauthentic behaviors in social media'' and PNRR MUR project PE0000013-FAIR.}

{\small
\bibliographystyle{ieee_fullname}

}

\end{document}